\documentclass[10pt,journal,compsoc]{IEEEtran}

\usepackage{graphicx}
\usepackage[noend]{algpseudocode}
\usepackage{epstopdf}
\usepackage[cmex10]{amsmath}
\usepackage{url}

\usepackage{microtype}

\newcommand\boldmu{\boldsymbol\mu}

\begin{document}

\markboth{U. Kamath et al.}{ Theoretical and Empirical Analysis of a Parallel Boosting Algorithm}

\title{ Theoretical and Empirical Analysis of a Parallel Boosting Algorithm}
\author{Uday~Kamath,~\IEEEmembership{Member,~IEEE,}
        Carlotta~Domeniconi,~\IEEEmembership{Member,~IEEE,}
        and~Kenneth~De Jong,~\IEEEmembership{Life~Fellow,~IEEE}
\IEEEcompsocitemizethanks{\IEEEcompsocthanksitem U. Kamath  is with the Ontolabs, Ashburn, VA 20147.\protect\\
E-mail: kamathuday@gmail.com
\IEEEcompsocthanksitem C. Domeniconi and  K. De Jong are with George Mason University.\protect\\
E-mail: carlotta@cs.gmu.edu, kdejong@gmu.edu}
\thanks{Manuscript received Febraury 11, 2015}}

\IEEEtitleabstractindextext{%
\begin{abstract}
Many real-world problems involve massive amounts of data.  Under these circumstances learning algorithms often become prohibitively expensive, making scalability a pressing  issue to be addressed. A common approach is to perform sampling to reduce the size of the dataset and enable efficient learning.  Alternatively, one customizes learning algorithms to achieve scalability. In either case, the key challenge is to obtain algorithmic efficiency without compromising the quality of the results.
In this paper we discuss a meta-learning algorithm (PSBML) which combines features of parallel algorithms with concepts from ensemble and boosting methodologies to achieve the desired scalability property. We present both theoretical and empirical analyses which show that PSBML preserves a critical property of boosting, specifically, convergence to a distribution centered around the margin.
We then present additional empirical analyses showing that this meta-level algorithm provides a general and effective framework that can be used in combination with a variety of learning classifiers. We perform extensive experiments to investigate the tradeoff achieved between scalability and accuracy, and robustness to noise, on both synthetic and real-world data. These empirical results corroborate our theoretical analysis, and demonstrate the potential of PSBML in achieving scalability without sacrificing accuracy.
\end{abstract}

}

\maketitle

\section{Introduction}
Many real-world applications, such as web mining, social network analysis, and bioinformatics, involve massive amounts of data.  Under these circumstances many traditional supervised  learning algorithms often become prohibitively expensive, making scalability a pressing  issue to be addressed.
For example,  support vector machines (SVMs) have training times of $O(n^3)$ and space complexity of $O(n^2)$, where $n$ is the size of the training set~\cite{Chang}.  

In order to handle the ``big data problem'', one of two approaches is typically taken: (1) Sampling of the data to reduce its size, or (2) customization of the learning algorithm to improve the running time via  parallelization.  Sampling techniques often introduce unintended biases that reduce the accuracy of the results.  Similar reductions in accuracy often result from modifications to a learning algorithm to improve its speed. The second approach also lacks generality, and requires customization per learning algorithm. As such, it is highly desirable to obtain  a general framework that can enable scalable machine learning with the following properties: (1) to be  applicable to a large variety of algorithms; and (2) to keep high accuracy while achieving the desired speed-up. This paper addresses this need, while focusing on classification.


Recently, a parallel spatial boosting machine learner (PSBML) was introduced~\cite{psbml,KamathDJ13}. 
PSBML is a meta-learning paradigm that leverages concepts from stochastic optimization and ensemble learning to arrange a collection of classifiers in a two-dimensional toroidal grid. 
The key novelty and relevance of this framework is how scalability is achieved and the fact that it overcomes the major limitations of existing approaches to big data, i.e., sampling and customized parallel solutions. Furthermore, due to its distributed nature, the paradigm is especially effective with massive data.
In this paper we provide both a theoretical and an empirical analysis of PSBML to investigate its behavior and properties. Our findings reveal that the significance of PSBML is two-fold.  First, it provides a general framework for parallelization, and as such it can be used in combination with a variety of learners. Second, it is capable of achieving effective speed-ups without sacrificing accuracy.

We use Gaussian mixture models (GMMs) combined with the mean-shift procedure to establish an analytical model of PSBML, and show that it  converges to a data distribution whose modes are centered on the margin of the classification boundary. As such, the algorithm inherits the properties of good generalization and resilience to noise that are associated with large margin classifiers. We perform extensive experiments to evaluate the performance of PSBML with a variety of learners, to measure the speed and accuracy trade-offs achieved, and to test the effect of noise. All results confirm the strength of PSBML anticipated by our theoretical findings.

This paper is a major and significant extension of our previous work~\cite{psbml,KamathDJ13}. In particular, 
the novel contribution of this article is as follows. 
\begin{itemize}
  \item Formal proof showing the convergence property of the PSBML algorithm and the fact that PSBML  is a large margin classifier.
  \item Extensive empirical experiments as detailed below.
\begin{itemize}
	\item Experiments using simulated data to verify the established theoretical findings and properties of PSBML. The results confirm the expected behavior as anticipated by the theory developed in this paper.
\item Extensive experiments on scalability comparing nine different classifiers custom optimized for speed, including different versions of support vector machines and parallel boosting.

\item Experiments on PSBML scalability as a function of data size and number of threads. Our findings show that training time scales linearly with data size and it steadily improves with the number of threads. 

\item Experiments of PSBML memory requirement as a function of data size. Our findings show a linear increase of the mean peak working memory with the training data size. 

	\item Additional experiments on parameter sensitivity analysis, meta-learning, and noise impact, thus providing a comprehensive testing framework. 
\end{itemize}

\end{itemize}


\section{Related Work}
\label{relatedwork}
The PSBML approach builds upon  stochastic optimization techniques and ensemble learning. The specific stochastic optimization technique used is spatially structured  evolutionary algorithms, which embed the data  in a metric space which constrains how samples may interact, and how are compared and updated~\cite{Sarma96ananalysis,alba05}.  In ensemble learning, multiple classifiers are generated and combined to make a final prediction. It has been shown that ensemble learning, through the consolidation of different predictors, can lead to significant reductions in generalization error~\cite{Bennett02exploitingunlabeled}.
Of particular relevance is the Adaboost technique and its variants, such as confidence-based boosting~\cite{Schapire99improvedboosting}. Adaboost induces a classification model by estimating the hard-to-learn instances in the training data~\cite{FreundS96}.
A formal analysis of the AdaBoost technique has derived theoretical bounds on the margin distribution to which the approach converges~\cite{Schapire97boostingmargin}.

In statistical learning theory, a formal relationship between the notion of {\textit{margin}} and the generalization classification error has been established~\cite{Vapnik:1995}. As a result, classifiers that converge to a large margin perform well in terms of generalization error. One of the most popular examples of such classifiers is support vector machines (SVMs). The classification boundary provided by an SVM has the largest distance from the closest training point. SVMs have been modified to scale to large data sets~\cite{liblinear,fm:02b,bottou-bousquet-2008,Tsang:2007,Joachims:1999}.   Many of these adaptations introduce a bias caused by the used approximation, like sampling the data or assuming a linear model, that can lead to a loss in generalization while trying to achieve speed.

To achieve scalable solutions with large data, algorithm-specific customizations are performed to enable distributed architectures and network computing~\cite{psvm,FouDroDunEasGosIngManOweWet10,NIPS2006725,WoodsendG09}. These modifications have been conducted on algorithms like decision trees, rule inductions, and boosting algorithms~\cite{Shafer96sprint,tree2,ParallelBoostingLazarevic,ParallelBoostingTKDE}. In most cases, the underlying algorithm needs to be changed in order to achieve a parallel computation. Use of MapReduce with machine learning is a notable demonstration of this difficulty. MapReduce is a popular method for a divide-and-conquer-based parallelism and has been used in conjunction with machine learning algorithms to scale to large datasets~\cite{Dean:2008,NIPS2006725}. Distributed machine learning systems such as Mahout or Pegasus~\cite{mahout,pegasus} sit on top of Hadoop, a common MapReduce implementation \cite{hadoop}. Many of the traditional machine learning algorithms need either significant per-algorithm customization, or must approximated to fit into the MapReduce framework.

MapReduce based algorithms have the additional disadvantage of unnecessary data movement and inefficiencies in iterative computation, which is a core part of most machine learning algorithms~\cite{lasalle2013mpi}. Recent proposed alternatives, including modifications of MapReduce (for example \cite{haloop,huang2011scalable,DBLP:journals/corr/abs-1204-6078}) can improve this situation significantly for highly iterative scenarios, reducing data movement by up to 1000 times in some cases. These methods typically have the same disadvantage as MapReduce with regard to algorithmic customization.
Finally the possible presence of heterogeneous nodes in clusters or cloud-based systems (with machines differing in terms of number of cores, RAM, and disk size) presents a challenge to such techniques. It can be difficult to achieve efficient utilization of heterogeneous node resources in divide-and-conquer methods resulting in either a poor utilization of resources or poor performance~\cite{lasalle2013mpi,rao2012performance,xie2010improving}.

We argue that what is needed is a generic framework which can efficiently be deployed to a variety of machine learning algorithms, and still efficiently uses heterogeneous networks of nodes. The logical spatial grid of learners we discuss in this paper promises to achieve exactly this.


In this paper, to analyse the PSBML algorithm we make use of Gaussian Mixture Models (GMMs) and the mean-shift procedure. A GMM is a parametric probabilistic model consisting of a linear combination of Gaussian distributions with unknown parameters.  Typically the parameter values are estimated so that the resulting model is the one that best fits the data~\cite{Rasmussen00theinfinite}.

Mean-shift  is a local search algorithm whose aim is to find the modes (i.e. local maxima) of a  distribution. It achieves this goal by performing kernel density estimation, and iteratively locating the local maxima of the kernel mixture as the zeros of the corresponding gradient function~\cite{Carreira03}. Convergence to local maxima is guaranteed from any starting point. Furthermore, it has been shown that, when combined with GMMs, mean-shift is equivalent to an expectation-maximization (EM) algorithm~\cite{Carreira05}.
The key advantage of using the mean-shift algorithm for mode finding on a given density is two-fold: (1) the approach is deterministic and non-parametric, since it is based on kernel density estimation; and (2) it poses no \emph{a priori} assumptions on the number of modes~\cite{Carreira03,Carreira05}. 

\section{The PSBML Algorithm}
\label{PSBMLalgorithm}

PSBML can be described as a meta-learning algorithm that  arranges a collection of learners in a two-dimensional toroidal grid. It can use any classifier capable of producing confidence measures on predictions. Each learner works independently on a portion of the data, and shares its ``results'' only with its neighbor learners. Through this local interaction, the information discovered locally at each cell (i.e., learner) gradually travels throughout the entire grid. Eventually, the collaborative behavior of the learners enables the emergence of the crucial information to solve the problem at hand. The emerging behavior only requires local interaction among the learners, thus enabling a high degree of parallelism. 

In the following, we  describe the different phases of the algorithm.
The pseudo-code of PSBML is  given in Algorithm \ref{pseudo-code}. 

\subsection{Initialization}

Given a collection of labeled data, an independent fixed validation set is created and the rest is used for training. PSBML uses the concept of wrap-around toroidal  grid to distribute the training data and the specified classifier to each node in the grid. The training data is distributed across all the nodes using stratified uniform sampling of the class labels (Line 1 of Algorithm \ref{pseudo-code}).
The parameters for grid configuration, i.e. width and height of the grid, replacement probability, and maximum number of iterations, are all included in {\it{GridParam}}.


\subsection{Node behavior at each epoch}
\begin{figure}
\centering
\includegraphics[width=0.9\columnwidth]{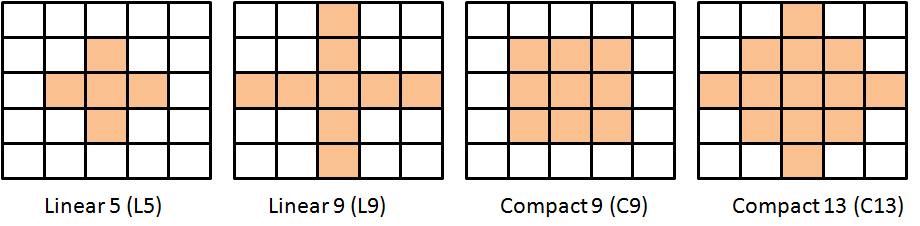}
\caption{Two dimensional grid with various neighborhood structures.}
\label{Grid}
\end{figure}

The algorithm runs for a pre-determined number of iterations (or epochs) (Line 4).
The behavior of a node at each epoch can be divided into two phases: training and testing. During training, a node performs a standard training process using its local data (Line 5).  For testing, a node's training data is combined with the instances assigned to the neighboring nodes (Lines 7 and 8).  Each node in the grid interacts only with its neighbors, based on commonly used neighborhood structures as shown in Figure \ref{Grid}. 
Each classifier outputs a confidence value for the prediction of each test instance, which is then used for weighting the corresponding instance. Every node updates its local training data for the successive training epoch by probabilistically selecting instances based on the assigned weights (Line 9).

The confidence values are used as a measure of how difficult it is to classify a given instance, allowing a node to select, during each iteration, the most difficult instances from all its neighbors.
Since each instance is a member of the neighborhood of multiple nodes, an ensemble assessment of difficulty is performed, 
similar to the boosting of the margin in AdaBoost~\cite{Schapire97boostingmargin}. Specifically, in PSBML the confidence $\text{\it cs}_i$ of an instance $i$  is set equal to the smallest confidence value obtained from any node and for any class:
$
\text{\it cs}_i = \min_{n \in N_i} c_{ni},
$
where  $N_i$ is a set of indices defined over the neighborhoods to which instance $i$ belongs, and $c_{ni}$ is the confidence credited to instance $i$ by the learner corresponding to neighborhood $n$.
These confidence values are then normalized through linear re-scaling:
$$
\text{\it cs}_i^{\text{\it norm}} = \frac{\text{\it cs}_i-\text{\it cs}_{\text{\it min}}}{\text{\it cs}_{\text{\it max}}-\text{\it cs}_{\text{\it min}}}
$$
where $\text{\it cs}_{\text{\it min}}$ and $\text{\it cs}_{\text{\it max}}$ are the smallest and the largest confidence values obtained across all the nodes, respectively.
The weight  $w_i = (1 - \text{\it cs}_i^{\text{\it norm}} )$ is assigned to instance $i$
to indicate its relative degree of classification difficulty.
The $w_i$s are used to define a probability distribution over the set of instances $i$, and used by a node to perform a stochastic sampling technique (i.e. weighted sampling with replacement) to update its local set of training instances.  The net effect is that, the smaller the confidence credited to an instance $i$ is (i.e. the harder it is to learn instance $i$), the larger the probability will be for instance $i$ to be selected. Instead of deterministically replacing the whole training data at a node with new instances, a  replacement probability $P_r$ is used. The effect of changing its value is discussed and analyzed in Section \ref{sensitivity}.
Due to the weighted sampling procedure, and to the constant training data size at each node, copies of highly weighted instances will be generated, and  low weighted instances will be removed with high probability during each epoch.

\subsection{Grid behavior at each epoch}
At each iteration, once all nodes have performed the local training, testing, and re-weighting, and have generated a new training dataset sampled from the previous epochs as described above, a global assessment of the grid is performed to track the ``best'' classifier throughout the entire iterative process.
The unique instances from all the nodes are collected and used to train a new classifier (Lines 10 and 11). The independent validation set created during initialization is then used to test the classifier (Line 12). This procedure resembles the ``pocket algorithm'' used in neural networks, which has shown to converge to the optimal solution~\cite{pocketalgorithm}. 
The estimated best classifier is given in output and used to make predictions for unseen test instances (Line 17).

\subsection{Iterative process}
The weighted sampling process and the interaction of neighboring nodes enable the hard instances to migrate throughout the various nodes, due to the wrap-around nature of the grid. 
The rate at which the instances migrate depends on the grid structure, and more importantly on the neighborhood size and shape. Thus, the grid topology of classifiers and the data distribution across the nodes provides the parallel execution, while the interaction between neighboring nodes and the confidence-based instance selection give the ensemble and boosting effects.


\section{Theoretical Analysis: PSBML is a Large Margin Classifier}
\label{theory}
We use Gaussian Mixture Models (GMMs) combined with the mean-shift algorithm to model the  behavior of PSBML.
Specifically, we formally show that PSBML, through the weighted sampling selection process, iteratively changes the data distribution, and converges to a distribution whose modes are centered around the margin, i.e. around the hardest points 
to classify.
This is an important milestone, as it shows that PSBML inherits the properties of good generalization and resilience to noise that are associated with large margin classifiers, and thus further strengthen the promise that our proposed framework will be an efficient and effective paradigm to perform scalable machine learning with massive data.

\newcommand\GETS{$\leftarrow$\ }
\newcommand\CALL[2]{{\textsc{#1}}({#2})}
\newcommand\TO{ {\bf to} }
\newcommand\COMMENT[1]{\vspace{0.5em}\Statex\Comment\textit{#1}}
\renewcommand\algorithmicindent{1em}

\newtheorem{algm}{\textbf{Algorithm}}

\begin{algm}{\sc PSBML}{(Train, Validation, GridParam)}
\rm
\label{pseudo-code}
\begin{algorithmic}[1]
	\State \CALL{InitializeGrid}{Train, GridParam}\Comment{{\small\it Distribute the instances over the nodes in grid}}
	\State currentMin \GETS 100
	\State Pr \GETS GridParam.pr\Comment{{\small\it Probability of replacement}}
	\For {i \GETS 0 \TO GridParam.iter}\Comment{{\small\it Train all nodes}}
		\State \CALL{TrainNodes}{GridParam}
		\State \CALL{TestAndWeighNodes}{GridParam}\Comment{{\small\it Collect neighborhood data and assign weights}}
		\State PrunedData \GETS \{\ \} 
		\For{ j \GETS 0 \TO GridParam.nodes}
			\State NeighborData \GETS \CALL{CollectNeighborData}{j} 
			\State NodeData \GETS NodeData $\cup$ NeighborData 
			\State ReplaceData \GETS \CALL{WeighSampling}{NodeData, Pr} 
			\State PrunedData \GETS \CALL{Unique}{ReplaceData}\Comment{{\small\it Unique keeps one copy of instances in set}}
		\EndFor
		\State ValClassifier \GETS createNew(GridParam.classifier) \Comment{{\small\it New classifier for validation}}
		\State error \GETS \CALL{Validate}{PrunedData,Validation,ValClassifier}
		\Statex \Comment{{\small\it Use validation set to track model learning}}
			\State currentMin \GETS error 
			\State bestClassifier \GETS ValClassifier
			\State marginData \GETS PrunedData 
	\EndFor
	\State \Return{bestClassifier, marginData}
\end{algorithmic}
\end{algm}

Each grid node in the PSBML algorithm, along with its neighborhood structure, represents a sample of the whole dataset, where each point is weighted according to how difficult it is to be classified. 
In our analysis, we fit a Gaussian mixture model on the weighted points, and apply the mean-shift procedure to locate the modes of the resulting distribution. We show that, throughout the iterations of PSBML, as more data closer to the boundary are being selected, the data distribution will grow higher modes centered around the margin. 
These modes will be the ones visited by the mean-shift procedure, irrespective of the starting point.

Since each node in the toroidal grid has the same behavior, they all fit a Gaussian mixture model on their respective neighborhood. By consolidating the micro-behavior of the mean-shift procedure at each node, we obtain an overall convergence to a distribution with peaks centered around the boundary. Our analysis below, and the empirical results in Section \ref{setting}, confirm this.

\subsection {Distribution of a node at time {\textbf{\textit{t}}} = 1}
After the completion of the first iteration of PSBML, each classifier in the grid has been trained with its own data, and is tested on the  instances of the neighbors, to which it  assigns confidence values. A common approach to assess the confidence of a prediction for an instance is to measure its distance from the estimated decision boundary: the smaller the distance, the smaller the confidence will be. The resulting weight values drive the probability for a point to be selected for the successive iterations. Below we use a Gaussian mixture to model this  process. 

Consider a Gaussian mixture density of M  components $p(\mathbf{x}) = \sum_{m = 1}^M {p(m)p(\mathbf{x}|m)}$,
where the $p(m)$ are the mixture proportions such that $p(m)>0$, $\forall m=1,\ldots ,M$, and 
$ \sum_{m = 1}^M {p(m) =1}$. Each mixture component is a Gaussian distribution in ${\mathbf{R}}^D$, i.e. $\mathbf{x}|m \sim \mathcal{N}_D(\mathbf{\boldmu} _m,\Sigma_m)$, where ${\mathbf{\boldmu} _m} = {{\rm E}_{p(\mathbf{x}|m)}}[\mathbf{x}]$  and ${\Sigma _m} = {{\rm E}_{p(\mathbf{x}|m)}}[ (\mathbf{x} - {\mathbf{\boldmu} _m}){(\mathbf{x} - {\mathbf{\boldmu} _m})^T}] $  are the mean and covariance matrix of the Gaussian component $m$.


Let us first consider a known result for the mean-shift procedure applied to Gaussian mixture models to find the modes of the distribution~\cite{Carreira03}. No closed-form solution exists to this problem, so numerical iterative approaches have been developed. In particular, the fixed-point iterative method gives the following fixed-point solution~\cite{Carreira03}: $\mathbf{x}^{(t+1)} =  \mathbf{f}(\mathbf{x}^{(t)})$ where
\begin{equation}
\mathbf{x}\!=\!\mathbf{f}(\mathbf{x})\!= 
	\!\left(\sum_{m = 1}^M p(m|\mathbf{x})\Sigma_m^{-1}\!\right)^{\!\!\! -1}
	\!\sum_{m = 1}^M p(m|\mathbf{x})\Sigma_m^{ - 1}{\mathbf{\boldmu} _m}\!\!\!
\label{eq:gmm}
\end{equation}

Let us assume now that we model the sample data assigned to a node and to its neighbors using a Gaussian mixture distribution of M  
components in ${\mathbf{R}}{ ^{\rm D}}$. In our analysis, we consider only the distribution of one class; the argument stays the same for the other class due to the symmetry with respect to the boundary. We need to embed the weighted sampling process performed by PSBML in our Gaussian mixture modeling. 
Lets assume the optimal boundary between classes is known.
 Let $\mathbf{s} \in {\mathbf{R}}^D$ be a point on the boundary. We estimate the distance of a point ${\mathbf{x}}$ from the boundary by considering its distance from ${\mathbf{s}}$. 
At each iteration of the PSBML algorithm, the weights bias the sampling  towards those points which are closer to the boundary: the larger the weight of a point is, the larger is the probability of being selected. To embed this mechanism in the Gaussian mixture modeling, we set the $m^{th}$ component to be ${p}^\prime(\mathbf{x}|m) = w(\mathbf{x})*p(\mathbf{x}|m)$, where $w(\mathbf{x})$ is a Gaussian weghting function centered at ${\mathbf s}$: 
\begin{equation}
w(\mathbf{x}) = (2\pi)^{-D/2}|\Sigma_{\mathbf s} {|^{ - 1/2}}{e^{ - 1/2{{(\mathbf{x}- {\mathbf{s}})}^T}{\Sigma^{ - 1}_{\mathbf s}}(\mathbf{x} -  {\mathbf{s}} )}}
\nonumber
\end{equation}
and
\begin{equation}
p(\mathbf{x}|m) = (2\pi)^{-D/2}|\Sigma_{m} {|^{ - 1/2}}{e^{ - 1/2{{(\mathbf{x}- {\mathbf{\boldmu}_m })}^T}{\Sigma^{ - 1}_m}(\mathbf{x} -  {\mathbf{\boldmu}_m })}}
\nonumber
\end{equation}

We compute the gradient of ${p}^\prime(\mathbf{x}|m)$  with respect to the independent variable $\mathbf{x}$, while keeping the parameters $\mathbf{\boldmu}_m$ and $\Sigma_m$  fixed:
\begin{equation}
\begin{split}
\frac{{\partial p^{\prime}(\mathbf{x}|m)}}{{\partial \mathbf{x}}} = w(\mathbf{x})\frac{{\partial p(\mathbf{x}|m)}}{{\partial \mathbf{x}}} + p(\mathbf{x}|m)\frac{{\partial w(\mathbf{x})}}{{\partial \mathbf{x}}}
\label{gradient}
\end{split}
\end{equation}

Considering each derivative:
\begin{equation}
\frac{{\partial p(\mathbf{x}|m)}}{{\partial \mathbf{x}}} = p(\mathbf{x}|m){{\Sigma} ^{ - 1}_m}(\mathbf{\boldmu}_m  - \mathbf{x}) 
\nonumber
\end{equation}
\begin{equation}
\frac{{\partial w(\mathbf{x})}}{{\partial \mathbf{x}}} = w(\mathbf{x}){\Sigma} ^{ - 1}_{\mathbf{s}}(\mathbf{s}  - \mathbf{x})
\nonumber
\end{equation}
and substituting these results in equation (\ref{gradient}), we obtain:
\begin{equation}
\begin{split}
\frac{{\partial p^{\prime}(\mathbf{x}|m)}}{{\partial \mathbf{x}}}  = w(\mathbf{x})p(\mathbf{x}|m){{\Sigma} ^{ - 1}_m}(\mathbf{\boldmu}_m  - \mathbf{x}) 
&+ \\p(\mathbf{x}|m)w(\mathbf{x}){{\Sigma} ^{ - 1}_{\mathbf{s}}}(\mathbf{s}  - \mathbf{x})
\end{split}
\end{equation}

We now turn to the mixture of $M$ Gaussian distributions.
By the linearity property of the differential operator, we obtain:
\begin{equation}
\begin{split}
\nonumber
\frac{{\partial p(\mathbf{x})}}{{\partial \mathbf{x}}} =
  w(\mathbf{x})\sum\limits_{m = 1}^M {p(m)p(\mathbf{x}|m)\Sigma_m^{ - 1}}(\mathbf{\boldmu}_m  - {\mathbf{x}})+  \\
(\mathbf{s} - \mathbf{x}) w(\mathbf{x})\sum\limits_{m = 1}^M {p(m)p(\mathbf{x}|m)\Sigma_{\mathbf{s}}^{ - 1}}
\end{split}
\end{equation}

By setting the above gradient to $\mathbf 0$ and simplifying $w(\mathbf{x})$, we derive a fixed point iteration procedure that finds the modes of the distribution~\cite{Carreira03}:
\begin{equation}
\begin{split}
\nonumber
 \sum\limits_{m = 1}^M {p(m)p(\mathbf{x}|m)\Sigma_m^{ - 1}} (\mathbf{\boldmu} _m - \mathbf{x}) = \\
(\mathbf{x} - \mathbf{s})\sum\limits_{m = 1}^M {p(m)p(\mathbf{x}|m)\Sigma_{\mathbf{s}}^{ - 1}}
\end{split}
\end{equation}
Solving for $\mathbf{x}$, we obtain:
\begin{equation}
\mathbf{x} =   \frac{{\sum\limits_{m = 1}^M {p(m)p(\mathbf{x}|m)\Sigma _{\mathbf{s}}^{ - 1} \mathbf{s} + \sum\limits_{m = 1}^M {p(m)p(\mathbf{x}|m)\Sigma _m^{ - 1} \mathbf{\boldmu} _m } } }}{{\sum\limits_{m = 1}^M {p(m)p(\mathbf{x}|m)\Sigma _{\mathbf{s}}^{ - 1} + \sum\limits_{m = 1}^M {p(m)p(\mathbf{x}|m)\Sigma _m^{ - 1}} } }}
\nonumber
\end{equation}
Using the Bayes rule and simplifying $p(\mathbf{x})$:
\begin{equation}
\nonumber
\mathbf{x} =   \frac{{\sum\limits_{m = 1}^M {p(m|\mathbf{x})\Sigma _{\mathbf{s}}^{ - 1} \mathbf{s} + \sum\limits_{m = 1}^M {p(m|\mathbf{x})\Sigma _m^{ - 1} \mathbf{\boldmu} _m} } }}{{\sum\limits_{m = 1}^M {p(m|\mathbf{x})\Sigma_{\mathbf{s}}^{ - 1} + \sum\limits_{m = 1}^M {p(m|\mathbf{x})\Sigma _m^{ - 1}} } }}
\end{equation}
Rearranging, we obtain our fixed-point solution:
\begin{equation}
\begin{split}
\mathbf{x} =& \left(\sum_{m = 1}^M p(m|\mathbf{x})\Sigma _{\mathbf{s}}^{- 1} + \sum_{m = 1}^M p(m|\mathbf{x})\Sigma _m^{ - 1}\right)  ^{ - 1} \times \\
&\left(\sum_{m = 1}^M p(m|\mathbf{x})\Sigma _{\mathbf{s}}^{ - 1}\mathbf{s} + \sum_{m = 1}^M p(m|\mathbf{x})\Sigma _m^{ - 1}\mathbf{\boldmu} _m \right) 
\end{split}
\label{eq:psbml}
\end{equation}

Comparing equations~(\ref{eq:gmm}) and~(\ref{eq:psbml}) we can see that, by weighting the points according to their distance from the boundary, the modes of the resulting distribution become the weighted average of the means $\boldmu_m$ and ${\mathbf s}$.
That is, each local classifier, by assigning weights to points according to the confidence of the prediction, causes the modes to shift towards the points closest to the estimated boundary, i.e. towards its margin.

\subsection {Distribution of the Grid at time  {\large\textbf{\textit{t}}} = 1}
The whole grid itself is modeled as a Gaussian mixture (given by the collection of GMMs at each node). Thus, the same derivation given above, applied to the grid, shows that the overall data distribution will have the same modes emerging from the individual nodes, i.e. centered around the margin of the boundary.

\subsection {Final Distribution of the Grid}

After a number of iterations, at each node, data will be sampled according to the current distribution. 
We can show that all the nodes will converge to the same mode. Suppose that a node $i$, at time $t$, has a neighborhood with means $T(t)=\{\mathbf{\boldmu}_1^{(t)}, \ldots , \mathbf{\boldmu}_l^{(t)}\}$, and one of these means, say $\mathbf{\boldmu}_g^{(t)}$, is the closest (globally) to the boundary. During successive iterations, the sampling process causes the elimination of modes that are far from the boundary. Thus, after $k>0$ steps, the local distribution of node $i$ will have a smaller number of modes: 
$T(t+k)=\{\mathbf{\boldmu}_1^{(t+k)}, \ldots , \mathbf{\boldmu}_{l-m}^{(t+k)}\}$, with $l-m>0$. Due to the weighted sampling mechanism  (note that the sample size remains constant at each iteration), $\mathbf{\boldmu}_g^{(t)}=\mathbf{\boldmu}_g^{(t+k)} \in T(t+k)$.
The whole process converges when $T(t+1) =  T(t) $, or the mean shift is negligible, and at convergence $T(t)=\{\mathbf{\boldmu}_g^{(t)}\}$.
We observe that spatially structured replication-based evolutionary algorithms show a similar behavior, where the global best is spread deterministically across the nodes, until all the nodes in the grid converge to the same individual according to logistic takeover curves~\cite{sarma97}.

\section{Empirical Analysis of PSBML and GMM with Mean-shift}
\label{setting}
We performed a number of experiments to verify the established relationship between PSBML and GMMs with mean-shift.
We generated synthetic data on which we ran the  following experiments.

\begin{enumerate}

\item We ran  the PSBML algorithm using a $5\times5$ spatial grid with the C9 neighborhood (see Fig. \ref{Grid}) and a large margin classifier, and observed the population distribution change over the training epochs. 

\item We replaced each local classifier with a GMM with mean-shift, while  keeping the grid structure and neighborhood interaction unchanged. Each data instance is weighted \emph{a priori} using the Gaussian weighting function  as defined in the theoretical analysis. 
We ran GMM with mean-shift on each node and performed sampling iteratively at every training epoch exactly as in PSBML. We again observed the changes in the population distribution over time.
%

\item We removed the grid and ran GMMs with mean-shift estimation on the whole dataset, with each instance weighted according to its distance from the known boundary as above. We observed the data distribution and final modes at convergence, and compared them with those obtained in the previous setting. 
\end{enumerate}


 \begin{figure}[t]
     \centering
    \begin{tabular}{c}
 \includegraphics[width=.9\columnwidth]{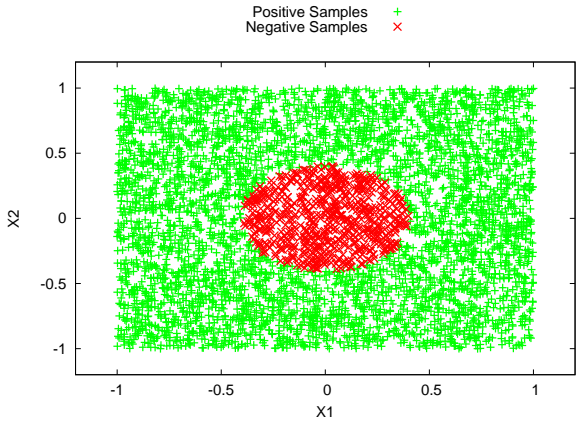}
      \end{tabular}
     \caption{Circle dataset.}
     \label{CircleAndGaussian} 
 \end{figure}

\begin{figure}
     \centering
    \begin{tabular}{c}
     \includegraphics[width=.9\columnwidth]{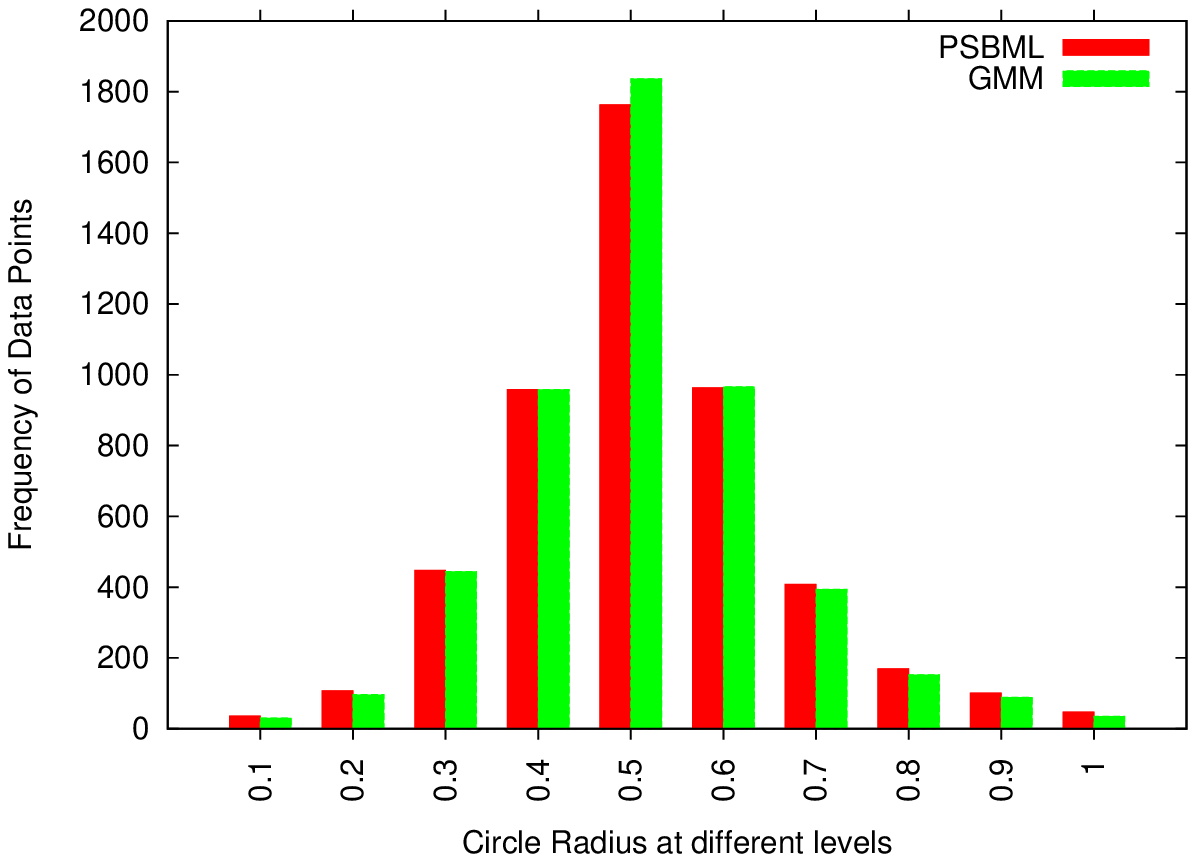} \\
     \includegraphics[width=.9\columnwidth]  {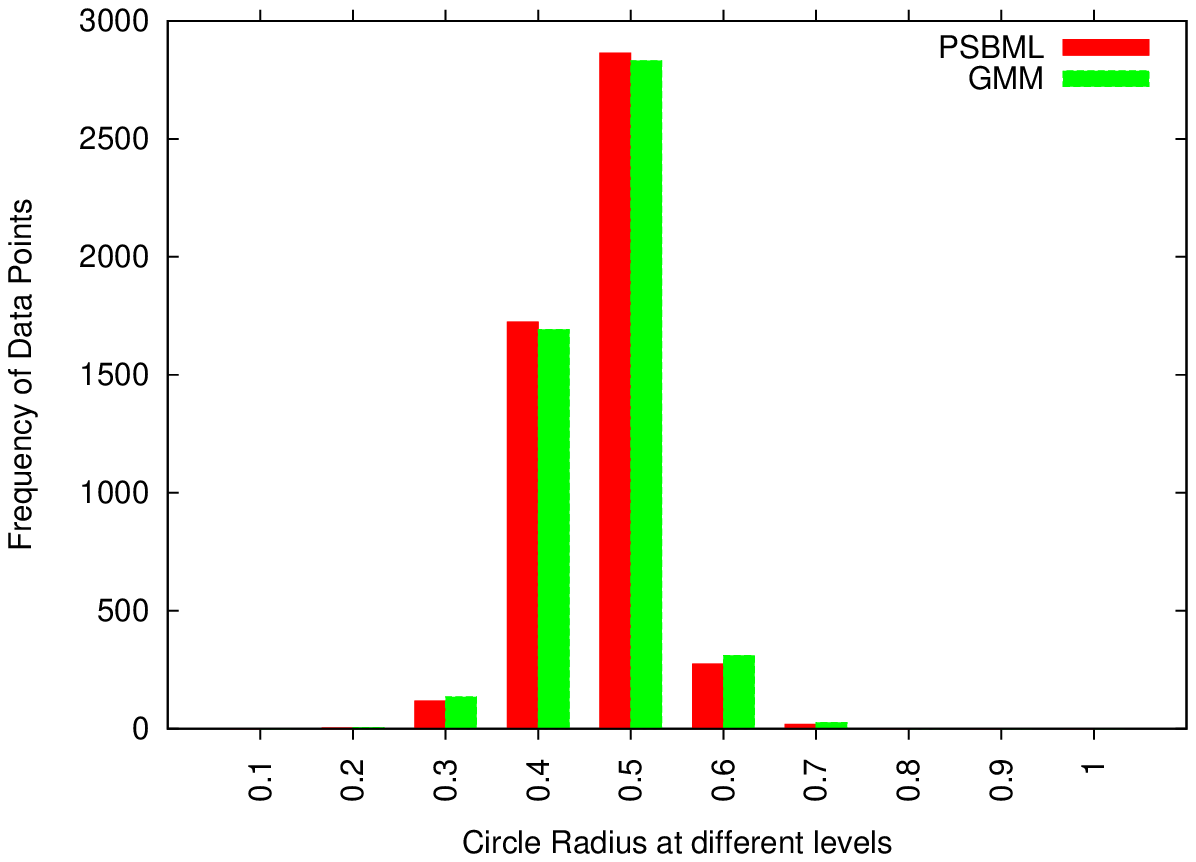}\\
     \end{tabular}
     \caption{Circle dataset: Data distribution at epochs 25 (Top) and 50 (Bottom) using PSBML and GMMs.}
     \label{Circle} 
 \end{figure}

\subsection {A Non-linearly Separable Dataset}
Instances were drawn at random within a square centered at the origin and with side of length two. Points with a distance smaller than 0.4 from the origin are labeled as negative, and those  with a distance greater or equal than 0.4 are labeled as positive (see Figure~\ref{CircleAndGaussian}). 
We ran the three experiments described in Section \ref{setting} on this data. For experiment 1, the large margin classifier used at each node fits a circle to its training set by setting its radius to the average distance of the origin from the smallest positive and the largest negative instances. For testing, the learner outputs ``$-$" when the instance falls within the circle, and ``$+$" otherwise. The confidence of the prediction is the distance of the instance from the circular boundary.

To compare the data distributions obtained in experiments 1 and 2, we recorded the number of points at various intervals of distances from the origin at training epochs 25 and 50. The resulting histograms are given  in Figure~\ref{Circle}. We can clearly observe that the two methodologies, PSBML and GMMs with mean-shift, provide a nearly identical distribution at both generations, and they converge to a distribution with modes centered on the points closest to the boundary. 

For experiment 3, we ran GMMs with mean-shift estimation 30 times  on the whole weighted data. The means of the modes at convergence were $(-0.01, 0.38)$  and $(0.01, -0.41)$,  with a very small standard deviation of 0.03. The distribution at convergence was very close to those obtained in experiments 1 and 2. Interestingly, we observed that, when the weights were removed, the modes at convergence moved to $(-0.03, 0.51)$ and $(0.03, -0.49)$. 

\subsection{Weight Distribution Changes} 
\begin{figure}[h]
     \centering
    \begin{tabular}{c}
     \includegraphics[width=.95\columnwidth]{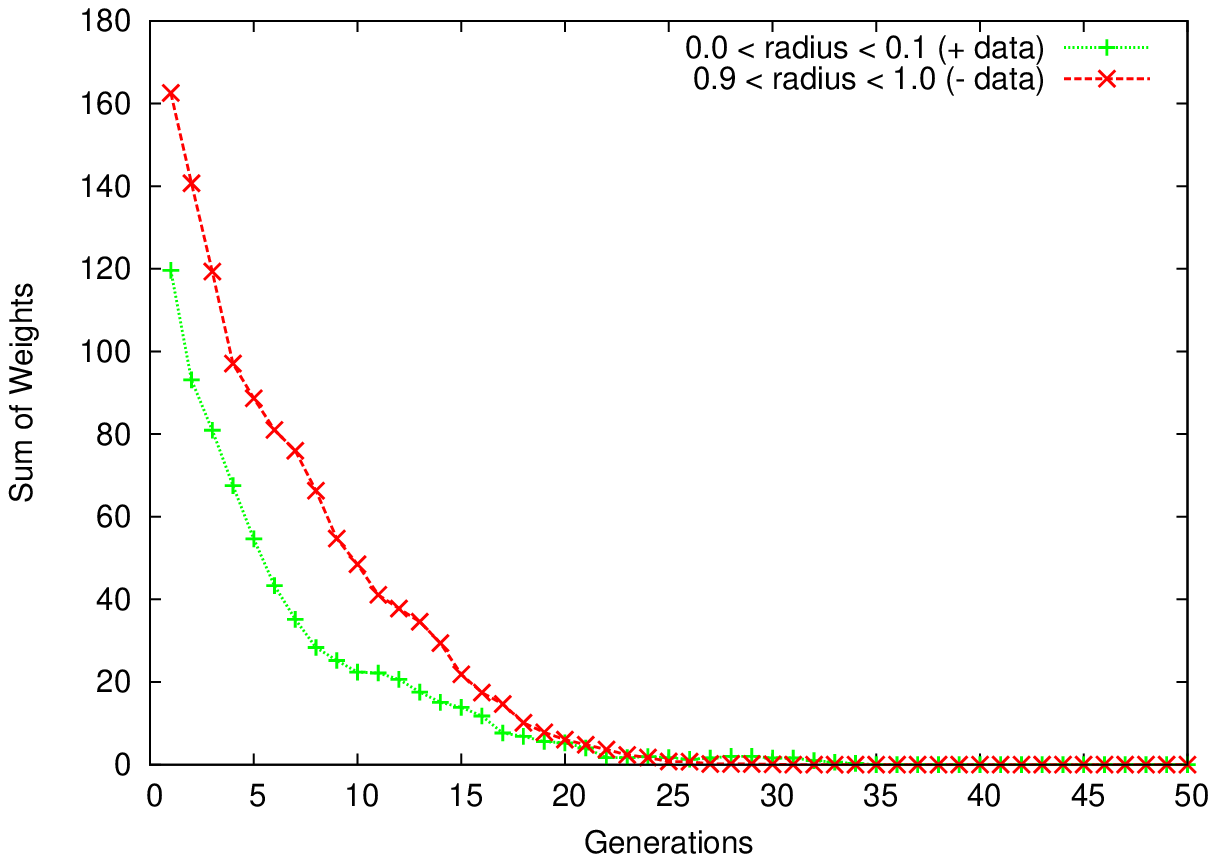} \\ 
     \includegraphics[width=.95\columnwidth]  {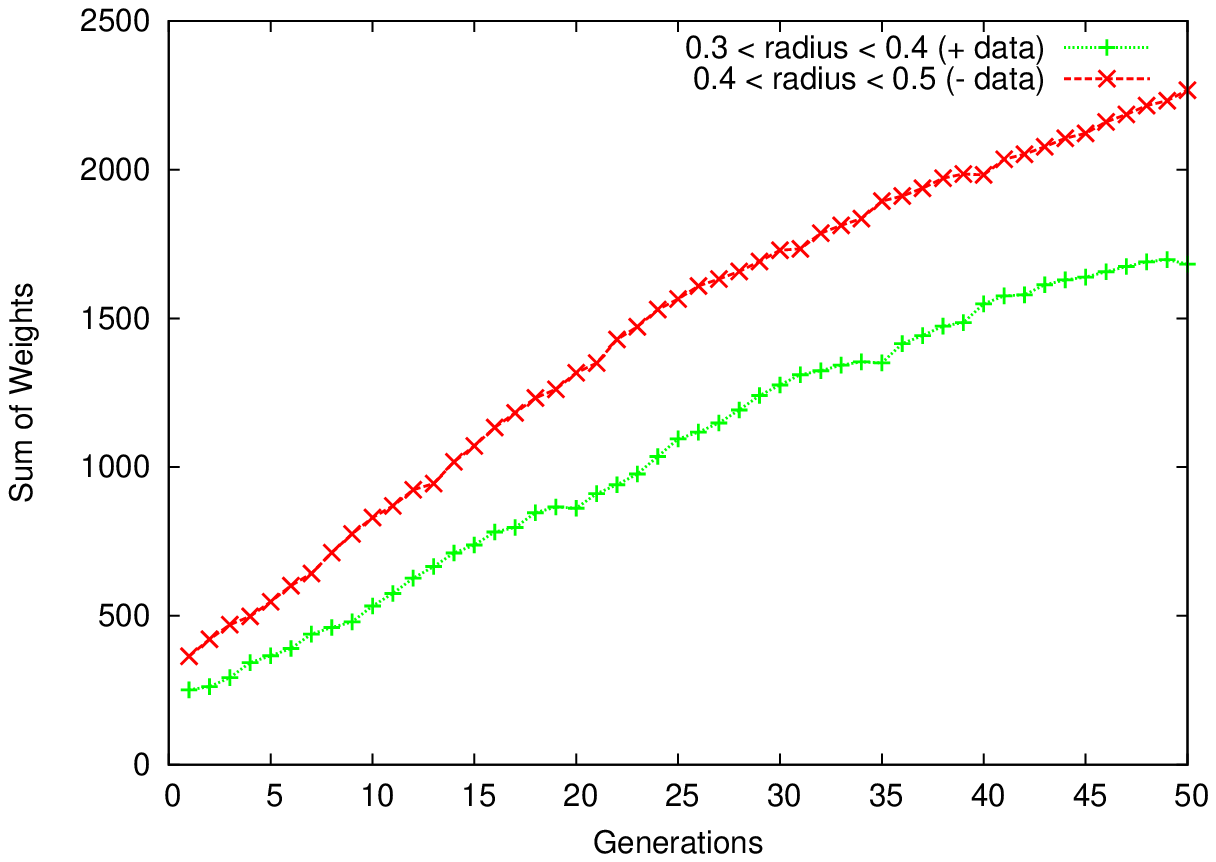}\\
     \end{tabular}
     \caption{Changes in weight distribution as function of time: (Top) exponential decay; (Bottom) logistic increase.}
     \label{WeightsPlot} 
 \end{figure}

One important property of boosting is to scale the weights of data as a function of its distance from the margin. 
To observe the effect of weight changes, in Figure ~\ref{WeightsPlot} we plotted the weights of all points at different radii and for different generations for the circle dataset (Figure \ref{CircleAndGaussian}). We can clearly see an exponential decay and a logistic increase based on the vicinity to the margin of the data. For positive points, when the radius is between 0.3 and 0.4, and for negative points, when the radius is between 0.4 and 0.5, an increase is seen with time, and for the rest there is an exponential decay, confirming a behavior analogous to boosting. 

\subsection{Linearly Separable Bivariate Gaussians}
\begin{figure}
 \begin{center}
    \begin{tabular}{cc}
\includegraphics[width=.4\textwidth]{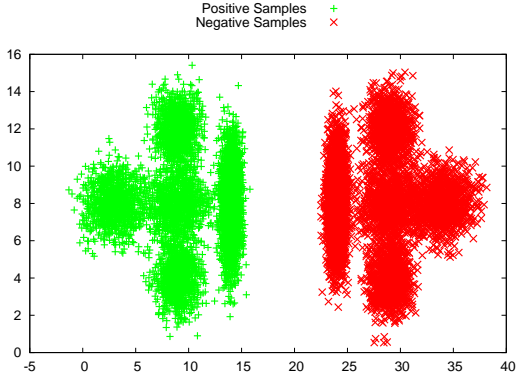} 
 \end{tabular}
     \caption{Bivariate Gaussian dataset.}
     \label{Gaussian} 
         \end{center}
 \end{figure}

We created a synthetic dataset consisting of 5 Gaussian distributions for each class, with roughly the same density but different shapes (see Figure~\ref{Gaussian}). The Gaussian distributions with means $(14,8)$ and $(24,8)$ are the closest to the boundary, given by the line $x = 20$. They simulate the ``global modes".
We again ran the three experiments described in Section \ref{setting}. The large margin classifier was simulated by estimating  the average distance between the smallest positive and the largest negative instances.

\begin{figure}
     \centering
    \begin{tabular}{c}
 \includegraphics[width=.9\columnwidth]{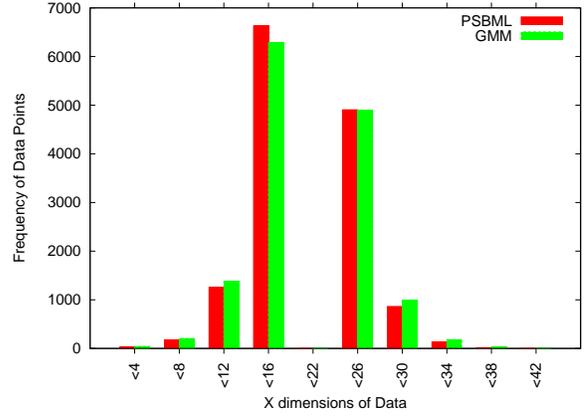} \\
\includegraphics[width=.9\columnwidth] {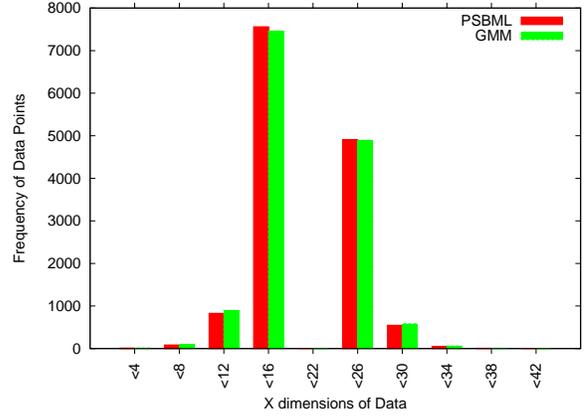}  \\
     \end{tabular}
     \caption{Linearly separable Gaussian dataset: Data distribution at epochs 25 (Top) and 50 (Bottom) using PSBML and GMMs.}
     \label{Linear} 
 \end{figure}

Again we observed that the data distributions produced by PSBML and GMM with mean-shift and grid structure are very much alike, as illustrated in Figure~\ref{Linear}. For experiment 3, with 30 runs on the weighted dataset, the modes of the data distribution converged to $(14.02, 7.89)$ and $(24.09, 7.88)$, with deviation of 0.002, matching exactly our results for experiments 1 and~2.

\subsection{Hard Instances and Support Vectors}
We also analyzed the data distribution at convergence by comparing the hard instances identified by PSBML with the support vectors of a trained SVM. Table~\ref{table:SupportVectorOverlap} shows the percentage of overlap for the two simulated datasets.
The support vectors of the trained SVMs with the highest $\alpha$ (i.e. weight) values  correspond to  the  hard instances with the top 10\% largest weights identified by the PSBML algorithm for both the datasets.
\begin{table}[h]
\begin{center}
\caption{Overlap percentage between support vectors and PSBML hard instances.\label{table:SupportVectorOverlap}}{
\begin{tabular}{@{}lrr@{}}
&\bf  2D Circle & \bf 2D Gaussians%
\vspace{0.1em}
\\
\hline
\vspace{-0.9em}\\
\it SV overlap   & 90\% & 94\% \\ 
\end{tabular}}
\end{center}
\end{table}

\section{Experimental Results}
\label{expresults}
We ran all scalability experiments (where running times were measured) on a dual 3.33 GHz 6-core Intel Xeon 5670 processor with no hyperthreading. This means that we had a maximum of 12 hardware threads available. PSBML was implemented both as a single threaded Weka~\cite{hall09} classifier and as a multithreaded standalone Java program that could run on any JVM version above 1.5 (see Section \ref{software}). All experiments with PSBML were run using a maximum heap size of 8GB and a number of threads equal to the number of nodes in the grid. All SVMs and boosting implementations,  where running times were compared, used either the native Matlab or C++ code, except for AdaBoostM1, where Weka 3.7.1 was used. All statistical significance tests were performed using the Matlab paired t-test function. 

\subsection{Parameter Sensitivity Analysis}
\label{sensitivity}
\begin{figure}
     \centering
    \begin{tabular}{c}
  \includegraphics[width=0.9\columnwidth]{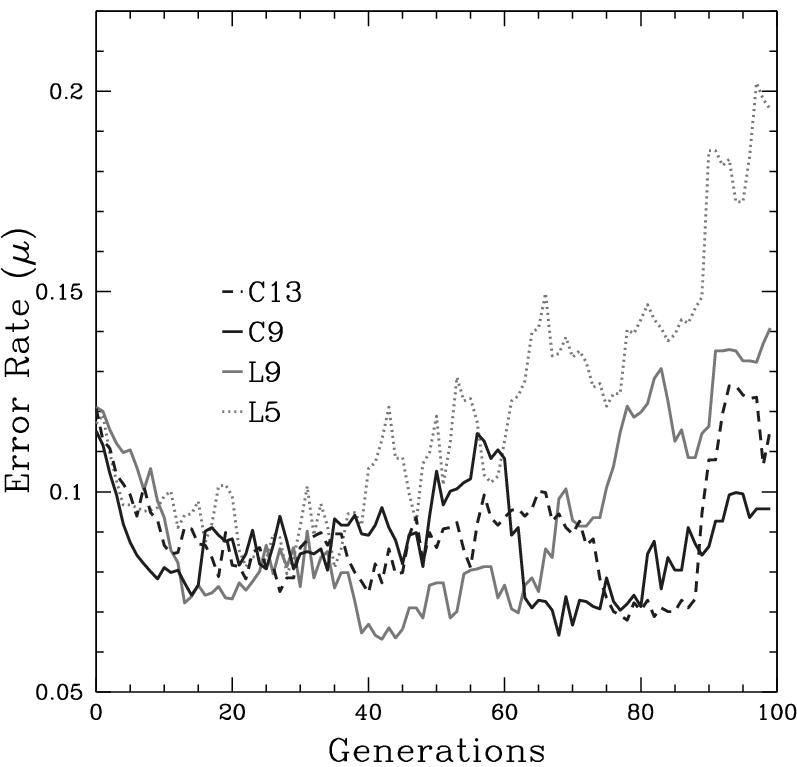}
  \end{tabular}
\caption{ Error rates at successive epochs
     for different neighborhood structures.}
\label{fig:ChessA}
\end{figure}

To study the effect that the neighborhood structure of the grid has on the performance of PSBML, we ran experiments on the
UCI Chess (King-Rook vs.\ King-Pawn) dataset, which consists of 3196 instances, 36 attributes, and 2 classes.
PSBML was run on this problem using various neighborhood structures, and the results are
shown in Fig.~\ref{fig:ChessA}. A $5 \times 5$ grid was used with a Na\"{\i}ve Bayes classifier with discretization for numeric features. PSBML was evaluated by combining the  instances selected by all the nodes at each epoch; using this collection of instances, we trained a single classifier and tested its performance on the test set.
Although the average size reduction of the training dataset was quite similar for all the neighborhoods, their classic ``over-fitting curves" were different (see Fig.~\ref{fig:ChessA}).  
The notion of {\it selection pressure} controlled by the parameter $P_r$ gives the degree to which only the highly weighted instances are selected at each epoch.  Since the sample selected at each node has a constant size, the selection pressure is driven by the size of the pool we choose the sample from. Furthermore, the more spread the neighborhood is, the faster the highly weighted instances travel through the grid. As such, the neighborhoods L9 and C13 have a stronger sampling pressure.
They produced a more rapid initial decrease in test classification error rates, which subsequently increased more rapidly as the training data became too sparse.  The simplest L5 neighborhood reduced classification error rates too slowly.  The best results were obtained with the neighborhood structure C9.

\begin{figure}
     \centering
    \begin{tabular}{c}
   \includegraphics[width=0.9\columnwidth]{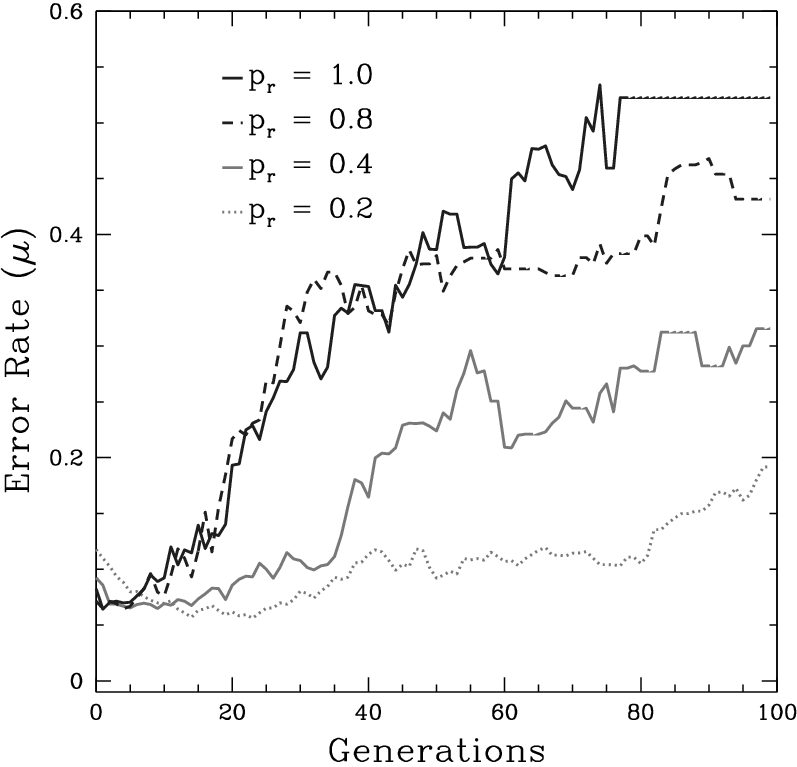}
  \end{tabular}
\caption{Error rates at successive epochs for different $P_r$ values. }
\label{fig:ChessB}
\end{figure}

\begin{figure}
     \centering
    \begin{tabular}{c}
   \includegraphics[width=0.9\columnwidth]{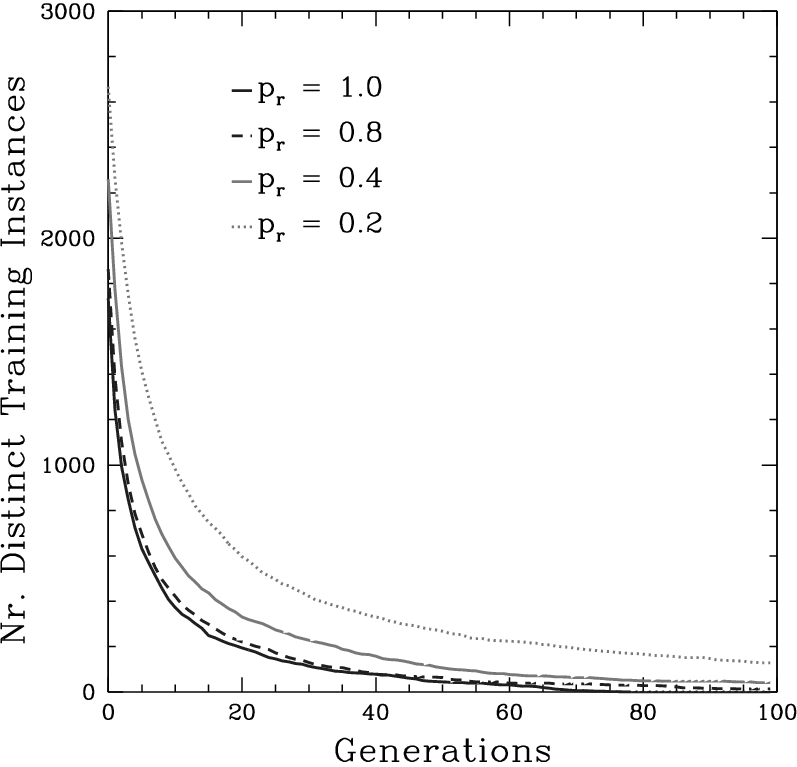}
    \end{tabular}
  \caption{Number of distinct instances sampled at successive epochs for different $P_r$ values.}
  \label{fig:ChessC}
\end{figure}

We used the UCI Chess dataset to also investigate how the rate of replacement $P_r$ affects the performance of PSBML. 
Figures~\ref{fig:ChessB} and~\ref{fig:ChessC}  illustrate that increasing the value of $P_r$ results in faster convergence rates, but also in less accurate models.
The best results were obtained when $P_r = 0.2$, which is the value we use in our experiments.

Finally, to investigate the impact of the grid size on accuracy, we used the Chess and Magik datasets, both with different training data sizes. The Magik dataset has 17,116 instances, 10 attributes, and 2 classes. We used the C9 neighborhood configuration and fixed the value of the replacement rate $P_{r}$ to $0.2$.  We tested various grid sizes ranging from $3 \times 3$ to $7 \times 7$. We measured the AUC for PSBML over $30$ runs. The Na\"{\i}ve Bayes classifier (with the same configuration) was used  at each node of the grid. Table \ref{table:gridsize} summarizes the results, showing that there is no statistically significant difference in AUC values across the various grid sizes.

These results are not surprising. Given the wraparound nature of the grid, and  the diffusion of hard instances through the weighted sampling process, only the rate of convergence to the margin is affected by the grid size. For the Chess dataset, which has only $3,196$ instances, as the number of nodes increases, we observe a slight degradation in performance. This is because the training data available at each node reduces significantly, and as a result the classifier's VC bound comes into effect~\cite{Vapnik:1995}. Thus, for smaller datasets, the choice of the grid configuration may depend on this lower bound. With a larger dataset like Magik ($17,116$ instances), no degradation is observed.  This is an important insight for the practitioner dealing with massive data, as scaling based on the number of hardware cores available can be used to configure the grid size.

\begin{table}[h]
\begin{center}
\caption{AUC results for PSBML with different grid sizes.\label{table:gridsize}}{
\begin{tabular}{cccccc}
Datasets & $3 \times 3$ & $4 \times 4$ & $5 \times 5$ & $6 \times 6$ & $7 \times 7$ \\
\hline
Chess & 98.5 & 98.5 & 98.3 & 98.2 & 98.1 \\
Magik & 89.4 & 89.5 & 89.4 & 89.4 & 89.5 \\
\end{tabular}}
\end{center}
\end{table}

\begin{table}%
\begin{center}
\caption{UCI datasets used in the experiments\label{table:Datasets}}{%
\begin{tabular}{@{}l@{~}rrrrr@{}}
&\bf Adult & \bf W8A & \bf ICJNN1   & \bf Cod & \bf Cover%
\vspace{0.1em}
\\
\hline
\vspace{-0.9em}\\
\it \# Train   &32560 & 49749 & 49990  & 331617 & 581012\\
\it \# Test    &16279 & 14951 & 91701 & 59535  & 58102 \\
\it \# Features   &123     & 300   & 22 & 8      & 54     \\
\it \# Labels   &2     &2   & 2  & 2      & 7    \\
\end{tabular}}
\end{center}
\end{table}%

\subsection{Meta-learning Experiments}
The goal of this experiment is two-fold: first to validate that PSBML provides a general framework for meta-learning, and therefore can be used in combination with a variety of learners; second, to verify that it's an effective parallel algorithm, i.e., it provides accuracy results comparable to the sequential counterpart, while achieving a speedup. To illustrate this, we performed experiments using three base classifiers: Naive Bayes, Decision Trees (C4.5), and Linear SVMs (LibLinear v1.8) (the corresponding Weka implementations were used). We used five medium to large UCI datasets~\cite{UCI}, commonly used for performance comparisons. Table \ref{table:Datasets} provides a description of the data. For each dataset, we normalized the features in the range [0,1], and converted multi-class problems to binary, using the one-vs-all strategy optimized for the LibSVM system, as described in~\cite{libsvm05}. The PSBML algorithm was run with the C9 neighborhood, a $3 \times 3$ grid, a replacement probability of 0.2, 20 training epochs, and a validation set size of $10\%$. We first optimized the base classifiers for performance, and then used the optimized settings in PSBML. Naive Bayes was used with the option of kernel estimation instead of using the default normal estimation; C4.5 was used with the default settings; and LibLinear was used with the $L_2$ loss function in both experiments. Each run, with the exception of Cover and C4.5, was repeated 30 times, and paired-t tests were used for statistical significance computation using the Area Under the Curve (AUC)~\cite{Bradley97theuse} as the metric. The experiments involving Cover and C4.5 were run only 10 times, due to the long processing time.  Hence significance is not recorded in this case. Results are reported in Table~\ref{table:metalearning1}. All statistically significant results are marked in bold-face. 

 \begin{table}
 \begin{center}
\caption{Meta-learning results (AUC) comparing the base classifiers and PSBML combined with the same.\label{table:metalearning1}}{%

\begin{tabular}{@{}l@{~~~}rr@{~~}rrr@{}}

 &\bf Adult & \bf W8A & \bf ICJNN1   & \bf Cod & \bf Cover%
\vspace{0.1em}
\\
\hline
\vspace{-0.9em}\\
NB   & 90.1 & 94.30 & 81.60  & 87.20 & 84.90  \\
PSBML    & \bf90.69 & \bf96.10 & \bf81.79 & \bf91.79  & \bf87.31%
\vspace{0.1em}
\\
\hline
\vspace{-0.9em}\\
C4.5   & 88.01 & \bf87.80 & 94.60  & 95.90  & 99.50\\
PSBML    & 88.78 & 84.80 & \bf97.30 & \bf97.24  & 97.44%
\vspace{0.1em}
\\
\hline
\vspace{-0.9em}\\
Linear SVM   & 54.60 & 80.20 & 64.60 & 88.80  & 72.20\\
PSBML    & \bf60.01 & 80.70 & 64.80 & \bf95.10  & \bf79.10 \\
\end{tabular}}
\end{center}
\end{table}

We observe that  PSBML, combined with the Naive Bayes classifier, performs statistically significantly better than the Naive Bayes classifier itself on all the datasets. Similar results were observed, and theoretical insights were provided, with regular boosting and Naive Bayes~\cite{elkan1997boosting}. Another important result to note is that the ensemble effect of PSBML makes the accuracy of a linear SVM significantly better (in three cases), while parallelizing the LibLinear SVM, which was already optimized for speed.

\subsection{Scalability Experiments}
The goal of this experiment is to validate whether PSBML performs competitively against custom optimized learning algorithms, in terms of training time, as a measure of speed, and in terms of accuracy, as a measure of performance. PSBML shares an important feature with SVMs: it reduces the  training data to the points which are close to the boundary. Thus, we  compared PSBML with a number of SVM implementations: a fast Newton based method, LP-SVM~\cite{fm:02b}, a  structural optimization-based technique, SVM-PERF~\cite{Joachims:1999} (linear because with an RBF kernel it crashed), the most commonly used LibSVM~\cite{libsvm05},  a fast optimized LibLinear~\cite{liblinear}, a stochastic gradient based approximation method, SGDT~\cite{bottou-bousquet-2008}, and fast ball enclosure-based BVM~\cite{Tsang:2007}. We also compared PSBML against a parallel AdaBoost algorithm~\cite{icsiboost} and the standard AdaBoostM1. All of the above mentioned implementations of SVMs incorporate some form of custom changes to boost the speed, like incremental  sampling of the dataset, or simplifying the quadratic optimization, or assuming linearly separable data.

\begin{figure}
\small
     \centering
    \begin{tabular}{cc}
     \includegraphics[width=.45\columnwidth]{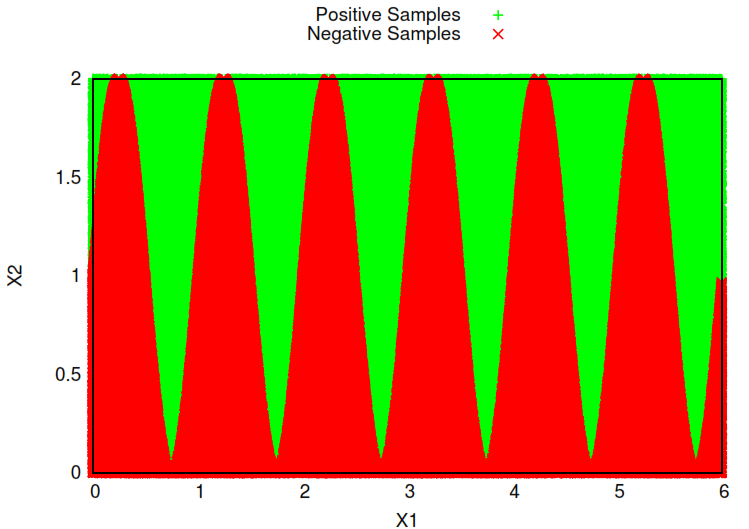} &
     \includegraphics[width=.45\columnwidth]  {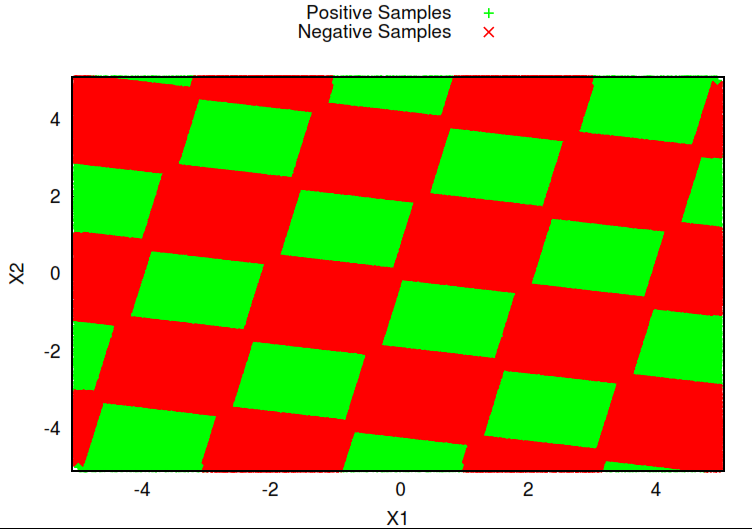}\\
     \end{tabular}
     \caption{Synthetic datasets: (Left) Sine wave; (Right) Checkerboard.}
     \label{Synthetic} 
 \end{figure}

The first dataset used for this experiment was a two dimensional decision boundary based on a sine wave generated by the function 
$f({\bf x}) = 2sin(2 \pi x_1)$ (see Figure~\ref{Synthetic}).
The dimension $x_1$ was sampled from the interval $[0,6.28]$ and the $y=f({\bf x})$ dimension was randomly sampled from the interval $[0,2]$.
The second dataset is a $4 \times 4$ rotated checkerboard data with alternate positive and negative classes as shown in Figure~\ref{Synthetic}. 
Each dataset has one million instances, and all the experiments were repeated 30 times. We measured training time for each of the runs, and the average training time is reported. 10 fold cross-validation was performed for accuracy and the average accuracy is reported. Each algorithm was tuned to some level of optimality for comparisons, i.e. the soft margin parameter and the radius of the RBF kernel for SVMs were optimized using a grid search in the intervals [-5,15] and [3,-15], respectively. 

The PSBML algorithm was run with the C9 neighborhood, a $3 \times 3$ grid, replacement probability of 0.2, 10 training epochs, and a validation set size of $10\%$ for each training fold. The C4.5 classifier with default parameters was used as it had an intermediate  training speed between the fast LibLinear and the kernel estimated Naive Bayes.
Results are shown in Table~\ref{table:CheckerboardAndSineWave}. For both the synthetic datasets, PSBML gives the most accurate results with respect to the methods that have comparable training speed (i.e., LibLinear and LibSVM). Most of the techniques customized for high speed give poor accuracy results.  The synthetic datasets, being highly non-linear, exaggerate the trade-offs implemented by the algorithms.

\begin{table}[h]
\begin{center}
\caption{Training speed (in seconds) and accuracy for the Checkerboard and the Sine Wave datasets.\label{table:CheckerboardAndSineWave}}{
\begin{tabular}{@{}l@{\hspace{-2em}\extracolsep{1em}}r@{~}r@{\extracolsep{1em}}r@{~}r@{}}
& \multicolumn{2}{c}{~\hspace{-3.25em}\bf \textit{Checkerboard}\hspace{-2.5em}~} & \multicolumn{2}{c}{\bf \textit{Sine Wave}}%
\vspace{0.1em}\\
\bf Algorithm & \bf Speed & \bf Acc& \bf Speed& \bf Acc%
\vspace{0.1em}
\\
\hline
\vspace{-0.9em}\\
{\it SVM} \\
\quad LP-SVM  (Linear) & 44.20 & 50.23  & 33.20 & 68.80\\
\quad LP-SVM (RBF) & 33.20 & 57.11 & 105.56 & 70.11\\
\quad LibLinear & 133.20 & 50.08 & 203.12 & 68.60\\
\quad SGDT (10 iterations) & 4.20  &54.49  & 4.20  &54.89\\
\quad SVM-PERF (Linear) & 1.10  & 51.01  & 2.01  & 61.90\\
\quad BVM (RBF) & 1.80 & 50.03 & 1.20 & 49.03\\
\quad LibSVM (RBF, $0.1\%$) & 136.20 & 98.20 & 423.23 & 70.80
\vspace{0.1em}\\
{\it Boosting} \\
\quad AdaBoostM1 &38.21 &51.25 &30.71 &74.25\\
\quad ParalleAdalBoost &17.90 &51.22 &13.90 &78.30\\
\qquad (9 threads,10 iterations)&&&&%
\vspace{0.3em}\\
{\it PSBML}&&&&\\
\quad PSBML (C4.5) & 123.10 &{\bf 99.49} & 193.10 & {\bf 99.56}\\

\end{tabular}}
\end{center}
\end{table}

\newcommand\ssp{\hspace{0.5em}}
\begin{table}[t]
\begin{center}
\caption{Training speed in secs, mis-classification, area under ROC and PRC for the KDD Cup 1999 dataset.\label{table:KDDCup99}}{
\begin{tabular}{@{}l@{\ssp}r@{\ssp}r@{\ssp}r@{\ssp}r@{}}
\bf Algorithm & \bf Speed & \bf MisClass & \bf ROC & \bf PRC%
\vspace{0.1em}
\\
\hline
\vspace{-0.9em}\\
{\it SVM} \\
\quad LibLinear & 80.20 & 25,447.3& 94.4& 6.3\\
\quad LibSVM (RBF, $1\%$)   & 90.20 & 25,517.8 & 94.1& 76.9\\
\quad LibSVM (RBF, $10\%$)  & 1,495.20 &25,366.1& 94.1 &13.1\\
\quad SGDT (10 iterations) & 211.10  &121,301 &-&-\\
\quad SVM-PERF (Linear) & 4.90  &25,877.1 & 93.1& 90.3\\
\quad BVM (RBF) & \bf3.20 & 25,451.3 &-&-%
\vspace{0.3em}\\
{\it Boosting} \\
\quad AdaBoostM1 &13,296.42  & 190,103.3 & 88.4 & 17.2\\
\quad ParallelAdaBoost & 202.30 & 26,170.2 & 36.2 &70.2\\
\multicolumn{5}{l}{\qquad(9 threads, 10 iterations)}%
\vspace{0.5em}\\
{\it PSBML}&&\\
\quad PSBML(C4.5) & 2,913.10 & \bf20,898.8& \bf95.6&91.2\\

\end{tabular}}
\end{center}
\end{table}

\subsubsection{Real-world Dataset}
The KDD Cup 1999 intrusion detection dataset was used to compare the performance of the algorithms. The dataset contains 4,898,431 training instances.
The problem was converted into a binary classification problem because many SVM implementations did not support multi-class labels. The feature set was also scaled within the range [0,1], which improved the performance of many SVMs almost 10 times. The PSBML algorithm was run with the C9 neighborhood, a $3 \times 3$ grid, replacement probability of 0.2, 10 training epochs, and a validation size of $0.1\%$ of the training data. C4.5 was used with default parameters again for the same reasons mentioned earlier.

\begin{figure}[t]
     \centering
     \includegraphics[scale=0.6]{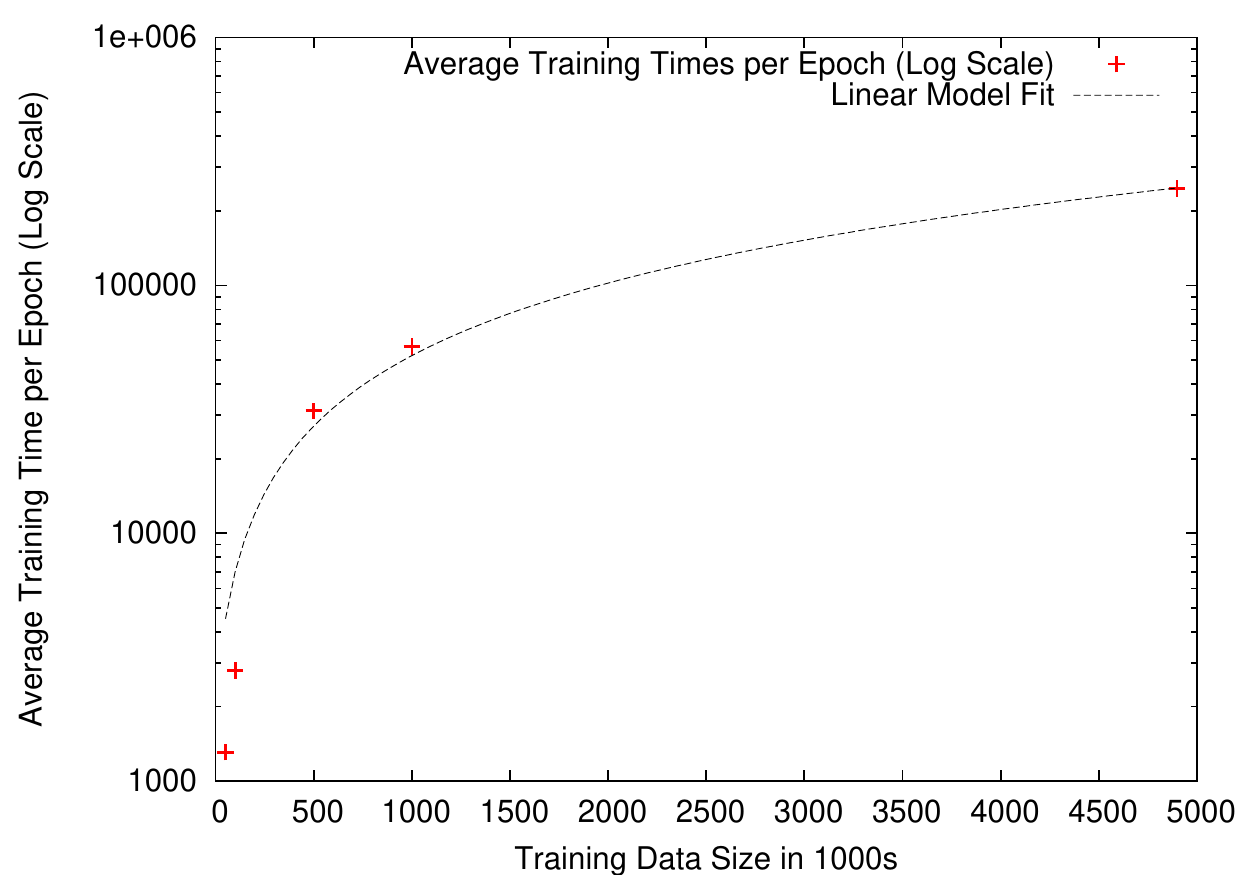}
\vspace{-0.5em}
     \caption{Mean training times per epoch with varying dataset sizes. }
     \label{Kdd99} 
 \end{figure}

\begin{figure}[t]
     \centering
     \includegraphics[scale=0.6]{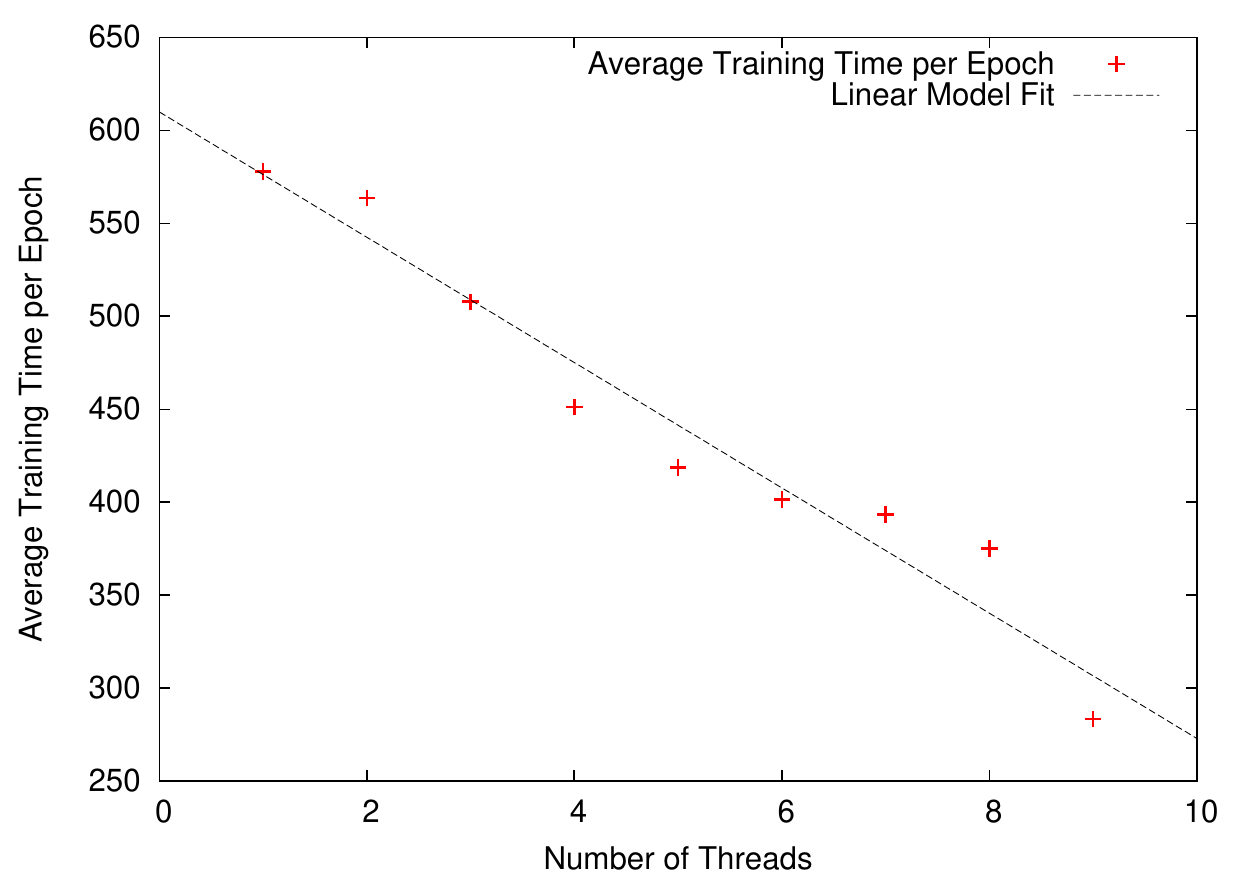}
\vspace{-0.5em}
     \caption{Mean training times per epoch with varying threads. }
     \label{Kdd99Threads} 
 \end{figure}

\begin{figure}[t]
     \centering
     \includegraphics[scale=0.6]{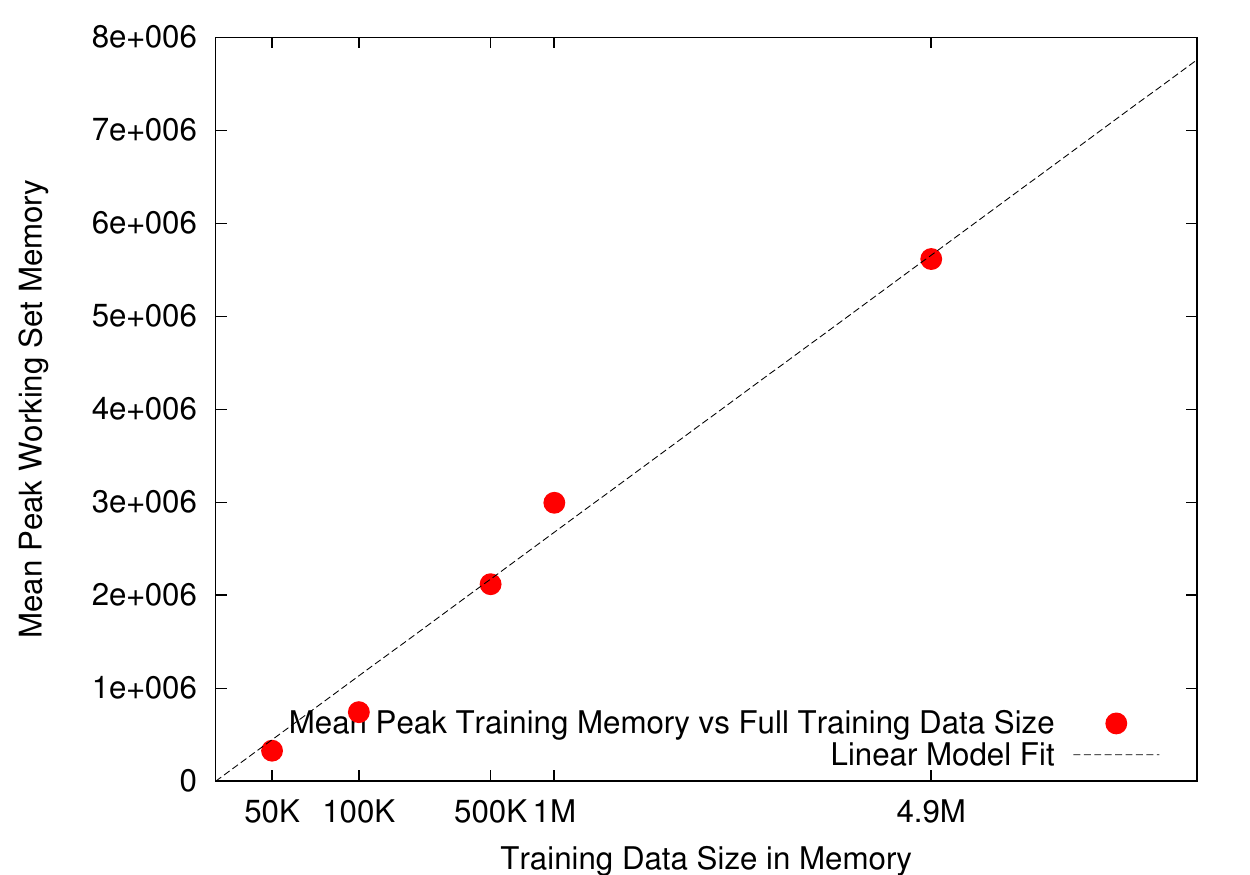}
\vspace{-0.5em}
     \caption{Mean peak working set memory  with varying dataset sizes. }
     \label{Kdd99Memory} 
 \end{figure}

In previous work, it was noted that many algorithms have a very similar error rate on this dataset. Hence, the number of mis-classifications was suggested and used as comparison metric~\cite{Yu:2003}. We do the same here. In addition, we measure the areas under the ROC and under the  Precision Recall Curve (PRC), since the dataset is unbalanced. Each of the experiments was run 30 times, except the AdaBoostM1 (only 10 times) due to large training time. The mean training times and the mean mis-classification averages are reported in Table~\ref{table:KDDCup99}. Some of the algorithms, e.g. LP-SVM, couldn't  run with a 12GB RAM machine, because the loading of the data matrix itself failed. Also, for  SGDT and BVM we couldn't compute the output probabilities to measure ROC and PRC due to the kernel choice. We observe that most algorithms that were optimized for speed had to trade-off accuracy. Also,  the training time of LibSVM increased considerably when the sampled data went from  $1\%$  to  $10\%$, with a small change in classification rate. 
The ROC value for PSBML was statistically significantly better; the value of the PRC area was comparable to that of SVM-PERF. In conclusion PSBML, while working on the entire dataset, finds a good classification rate at a considerable performance speed.

To see the impact of data sizes on PSBML, we also selected training samples of various sizes from $50K$, $100K$, $500K$, to one million. Ten runs were performed with standard PSBML with decision trees, a $3 \times 3$ grid,  and the C9 neighborhood. Nine threads were used in this experiment.
Training time (log scale) is plotted against data size in Figure~\ref{Kdd99}. The graph clearly shows a steady linear scaling with data size.
 To see the impact of the multi-core processor described above on scalability, we changed the number of threads and computed the corresponding average training times. The result is given in Figure~\ref{Kdd99Threads}, which shows again a consistent linear improvement with the number of threads.

Another important aspect of a large scale learning algorithm is memory requirements. To evaluate this impact,  we measured the memory usage with varying data sizes. We used the same data sizes and configuration as in the previous experiments. Figure~\ref{Kdd99Memory} shows the mean peak working memory during training as a function of different training data sizes. Again, this result shows a linear increase with the training data size, thus providing empirical evidence that  the memory space complexity of PSBML is  $O(n)$, where $n$ is the size of the training set. In comparison, SVMs are $O(n^2)$~\cite{Chang}.  As such, PSBML has a key advantage also in terms of memory requirement.

\subsection{Comparison against AdaBoost and Impact of Noise}
Here we compare  PSBML against AdaBoost and test the robustness in presence of noise. Previous work  found that boosting is more susceptible to noise as compared to other ensemble methods like bagging and stacking~\cite{Melville04experimentson,DBLP:conf/icml/LongS08}. We added noise to the class labels by randomly changing different percentages of labels. We used AdaBoostM1 both with decision stumps and with Naive Bayes (optimized using kernel estimators), and compared it against PSBML combined with the same underlying Naive Bayes classifier. PSBML was used with the default C9 neighborhood, replacement probability of 0.2, and validation set of  $10\%$.

We used the same datasets used for the meta-learning experiments, and did the same preprocessing. We performed 30 runs to compare the three algorithms without noise, and in presence of 10\% and 20\% of noise. The results are shown in Table \ref{table:noise}. Statistically significant results are highlighted in boldface.

In absence of noise, PSBML with Naive Bayes performs significantly better than AdaBoostM1 with decision stumps or with the same optimized Naive Bayes in three of the five datasets. To measure how robust a method is across all the datasets, we compute the following quantity:  $\text{\it impact}=  {1\over N}{{\sum_{i = i}^N ({{{ \overline{\text{\it auc}}}^ i}_{\text{\it no-noise}} - {{\overline{\text{\it auc}}}^i}_{\text{\it noise}}} }})$,
where $N$ is the number of datasets. The smaller the value of the impact is for an algorithm, the more robust that method is on average.

The impact values of AdaBoostM1 (DecisionStump), AdaBoostM1 (NaiveBayes), and PSBML (NaiveBayes) with 10\%  noise are 4.41, 3.32, and 1.71, respectively. Similarly, with 20\% noise the impact values for  
these algorithms
are 5.02,  4.62, and 2.02, respectively. This shows that the PSBML algorithm 
is more robust to noise as compared to standard boosting. This is likely due to two reasons. First, in PSBML, the weighted sampling procedure is driven by the confidence of predictions only (prediction errors are not used), while AdaBoost credits larger weights to instances which are erroneously predicted. Second, PSBML makes use of a validation set to estimate the best classifier to be used for prediction of test instances, thus preventing overfitting.

\begin{table}
\begin{center}
\caption{Performance of  AdaBoostM1 (DS: Decision Stump), AdaBoostM1 (NB: Naive Bayes) and PSBML (NB: Naive Bayes) with no, 10\%, and 20\% noise.\label{table:noise}} {
\begin{tabular}{@{}l@{~~}r@{~~}r@{~}r@{~~}r@{~~}r@{}}
&\bf Adult & \bf W8A & \bf ICJNN1   & \bf Cod & \bf Cover%
\vspace{0.1em}
\\
\hline
\vspace{-0.2em}\\
{\it \textbf{No Noise}}%
\vspace{0.1em}\\
AdaBoostM1/DS   & 87.10 & 77.80 & \bf93.40  & 92.80 & 75.70  \\
AdaBoostM1/NB & 87.20 & 93.30 & 84.30  &\bf95.70 & 85.30  \\
PSBML/NB    & \bf90.69 & \bf96.10 & 81.79 & 91.79  & \bf 87.31%
\vspace{0.4em}\\
{\it \textbf{10\% Noise}}%
\vspace{0.1em}\\
AdaBoostM1/DS   & 85.70 & 58.90 & \bf92.82 & 92.20  & 75.10\\
AdaBoostM1/NB   & 85.80 & 83.40 & 79.80  & \bf95.10  & 85.10\\
PSBML/NB    & \bf90.46 & \bf96.01 & 77.46 & 88.06  & \bf87.14%
\vspace{0.4em}\\
{\it \textbf{20\% Noise}}%
\vspace{0.1em}\\
AdaBoostM1/DS   & 85.10 & 57.10 &\bf 92.30  & 92.10  & 75.10\\
AdaBoostM1/NB   & 84.88 & 79.01 &79.70   & \bf94.90  & 84.20\\
PSBML/NB    & \bf90.10 & \bf95.97 &77.42  & 86.98  & \bf87.11\\

\end{tabular}}

\end{center}
\end{table}

\section{Conclusion}
\label{conclusions}
The PSBML algorithm provides a general framework for parallelizing machine learning algorithms.
The key contributions of this paper are: (1) Establishing a theoretical statistical model for PSBML, (2) Proving that PSBML is a large margin classifier, and (3) Providing a comprehensive experimental study. Our empirical analysis confirmed the veracity of the theoretical model. 

The meta-learning experiments have shown that PSBML exhibits characteristics similar to that of AdaBoost in the sense that adding ensemble boosting to a standard classifier produces at least comparable and often better results.
Scalability experiments confirm that while maintaining good running times for training, the accuracy is not compromised. We have also shown a steady linear improvement in speed with an increasing number of threads, as well as linear training time and linear memory use as a function of data size.
  In addition, the spatial structure aspects of PSBML provide a resilience to noise, an important feature for real-world applications.

There are several immediate extensions to this work.  We are now adapting the algorithm to semi-supervised learning and unsupervised learning. In addition, we are exploring the possibility of mapping PSBML onto distributed architectures like the  Beowulf-style clusters in combination with map-reduce algorithms.

\section{Software and Data}
\label{software}

Software, data, and parameters used to perform the experiments in this paper are available at https:/\!/sites.google.com/site/psbml2013/ under an academic license. 

\bibliographystyle{IEEEtran}
\bibliography{PSBML-2013_REFS}

\begin{IEEEbiography}[{\includegraphics[width=1in,height=1.25in,clip,keepaspectratio]{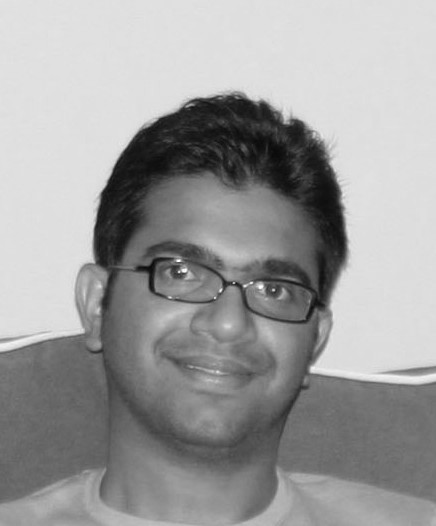}}]{Uday Kamath}

Dr. Uday Kamath received his Ph.D. in Information Technology from George
Mason University in 2014. He received his BS degree in electrical
electronics from Bombay University in 1996 and the M.S. degree in
computer science from the University of North Carolina at Charlotte in
1999.  He is the founder of Ontlolabs and has reserach interests in machine
learning, evolutionary algorithms, bioinformatics,  statistical modeling techniques and parallel algorithms. He is a member of the IEEE
and the ACM.

\end{IEEEbiography}

\begin{IEEEbiography}[{\includegraphics[width=1in,height=1.25in,clip,keepaspectratio]{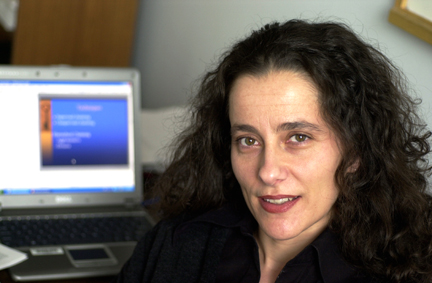}}]{Carlotta Domeniconi}
Carlotta Domeniconi is an Associate Professor in the Department of Computer Science at George Mason University. Her research interests include machine learning, pattern recognition, educational data mining, and feature relevance estimation, with applications in text mining and bioinformatics. She has published extensively in premier journals and conferences in machine learning and data mining. She was the program co-Chair of SDM in 2012 and the co-Chair of a number of workshops. Dr. Domeniconi has served as PC member and area chair for a variety of top-tier conferences; she was an Associate Editor of the IEEE Transactions of Neural Networks and Learning Systems Journal, and is currently serving on the editorial board of Knowledge and Information Systems and Computational Intelligence journals. Dr. Domeniconi is a recipient of an ORAU Ralph E. Powe Junior Faculty Enhancement Award. She has worked as PI or co-PI on projects supported by the NSF, the US Army, the Air Force, and the DoD, and she is a recipient of an NSF CAREER Award.
\end{IEEEbiography}

\begin{IEEEbiography}[{\includegraphics[width=1in,height=1.25in,clip,keepaspectratio]{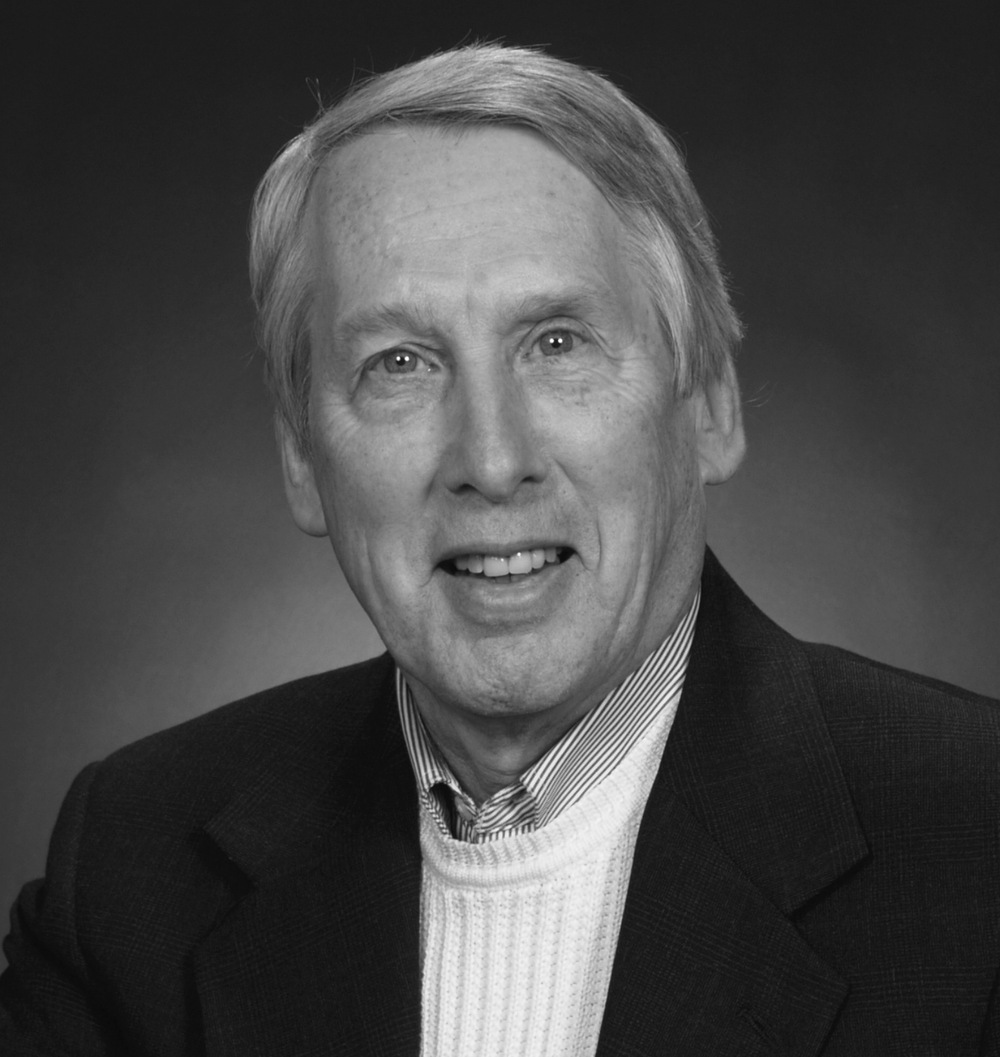}}]{Kenneth A. De Jong}
Dr. Kenneth A. De Jong is a professor of computer
science and an associate director of the
Krasnow Institute at George Mason University.
His research interests include evolutionary
computation, adaptive systems and machine
learning. He is an active member of the
evolutionary computation research community
with a variety of papers, PhD students, and
presentations in this area. He is also responsible
for many of the workshops and conferences on
evolutionary algorithms. He is the founding editor-in-chief of the journal
Evolutionary Computation (MIT Press), and a member of the board of
ACM SIGEVO. He is the recipient of an IEEE Pioneer award in the field
of evolutionary computation and a lifetime achievement award from the
Evolutionary Programming Society.
\end{IEEEbiography}

\end{document}